%% file: acl.tex
\pdfoutput=1

\documentclass[11pt]{article}

\usepackage[]{acl}

\usepackage{times}
\usepackage{latexsym}

\usepackage[T1]{fontenc}
\usepackage[utf8]{inputenc}
\usepackage{microtype}
\usepackage{graphicx}
\usepackage{booktabs}
\usepackage{blindtext}
\usepackage{graphicx}
\usepackage{capt-of}
\usepackage{booktabs}
\usepackage{varwidth}
\usepackage{multirow}
\usepackage{xspace,mfirstuc,tabulary}

\usepackage{amsmath}
\usepackage{amssymb}

\usepackage{subcaption}

\usepackage{pgfplots}
\usepackage{tikz}
\usepackage{hyperref}
\usetikzlibrary{patterns}
\usepackage{xcolor,colortbl}

\title{Towards General Text Embeddings with Multi-stage Contrastive Learning}

\author{
Zehan Li$^1$, Xin Zhang$^1$,
Yanzhao Zhang$^1$, Dingkun Long$^1$,
Pengjun Xie$^1$, Meishan Zhang \\
$^1$Alibaba Group \\
{\tt \{lizehan.lzh,linzhang.zx,zhangyanzhao.zyz,} \\
{\tt dingkun.ldk,pengjun.xpj\}@alibaba-inc.com}
}

\newcommand{\cmark}{\checkmark}

\begin{document}
\maketitle
\begin{abstract}
We present GTE, a general-purpose text embedding model trained with multi-stage contrastive learning.
In line with recent advancements in unifying various NLP tasks into a single format, we train a unified text embedding model by employing contrastive learning over a diverse mixture of datasets from multiple sources. By significantly increasing the number of training data during both unsupervised pre-training and supervised fine-tuning stages, we achieve substantial performance gains over existing embedding models.
Notably, even with a relatively modest parameter count of 110M, GTE$_\text{base}$ outperforms the black-box embedding API provided by OpenAI and even surpasses 10x larger text embedding models on the massive text embedding benchmark.
Furthermore, without additional fine-tuning on each programming language individually, our model outperforms previous best code retrievers of similar size by treating code as text.
In summary, our model achieves impressive results by effectively harnessing multi-stage contrastive learning, offering a powerful and efficient text embedding model with broad applicability across various NLP and code-related tasks.\footnote{The GTE model is publicly available at \url{https://huggingface.co/thenlper/gte-large}}
\end{abstract}

\input{sections/intro}

\input{sections/rw}

\input{sections/approach}

\input{sections/results}

\input{sections/analysis}

\input{sections/discussion}

\section{Conclusion}

This paper presents a multi-stage contrastive learning approach to develop text embedding model that can be applied to various tasks. Our model benefits from a diverse training data mixture, enabling it to achieve good generalization performance for single vector embedding. Through extensive evaluation on multiple benchmarks, we demonstrate the effectiveness and versatility of our text embedding model. Our future work will focus on scaling the model to support longer context, extending it to support multilingual and multi-modal applications, as well as exploring the benefits of prompts and instructions.


\bibliography{anthology,custom}
\bibliographystyle{acl_natbib}

\appendix
\input{sections/appendix}
\end{document}

%% file: sections/intro.tex
\section{Introduction}
Text embeddings have became an indispensable component in many natural language processing tasks, such as text classification, text retrieval, question answering and dialogue systems~\citep{karpukhin-etal-2020-dense,Humeau2020Poly-encoders:,Choi2021EvaluationOB,izacard2022unsupervised,Long2022MultiCPRAM,Rajapakse2023DensePR}. These embedding models represent texts using low-dimensional vectors and capture their similarity through vector operations. The emergence of recent large language models (LLMs)~\cite{gpt,Touvron2023LLaMAOA,OpenAI2023GPT4TR} has generated considerable interest in retrieval-augmented systems based on text embedding models that integrate the reasoning and comprehension capabilities of LLMs~\cite{izacard_few-shot_2022,Ram2023InContextRL,Shi2023REPLUGRB}. Consequently, there has been a growing focus on general text representation in both industry and academia. 

\begin{figure}[t]
    \centering
    \includegraphics[width=0.48\textwidth]{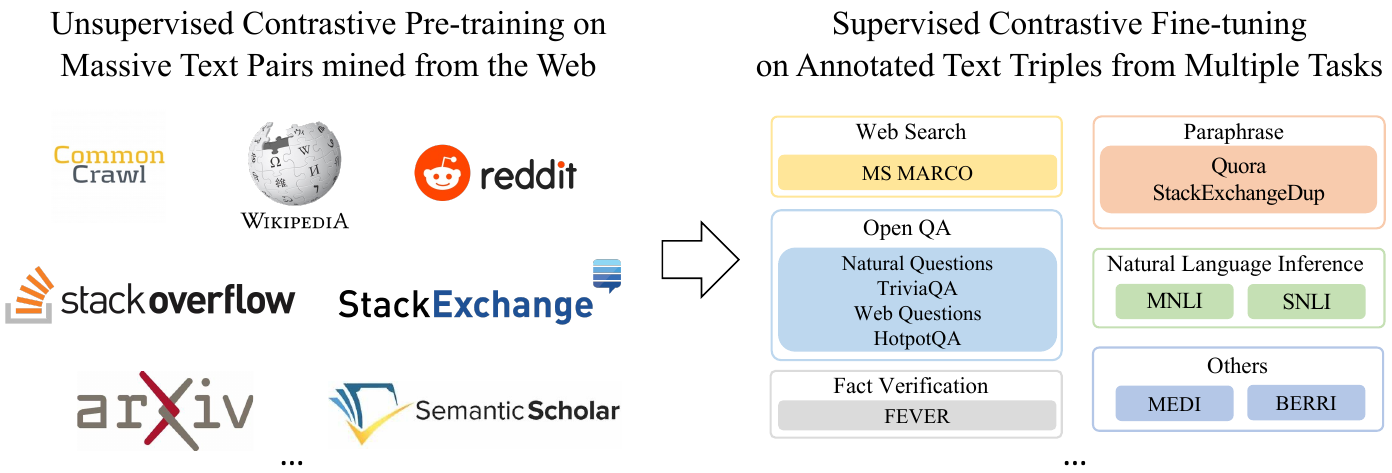}
    \caption{Illustration of the multi-stage contrastive learning pipeline used to train our text embedding model.}
    \label{fig:training}
\end{figure}

The pursuit of developing a unified model to address a multitude of downstream tasks has been long-standing due to the diverse formats, domains and downstream applications of natural language. The emergence of pre-trained language models has further opened up possibilities for training such a universal model. Nonetheless, within the realm of text representation research, previous text embedding models have primarily focused on specific tasks, and their training strategies or models, tailored to a single task, may not perform optimally in other contexts. For example, the text representation model SimCSE~\cite{gao-etal-2021-simcse}, trained on symmetric text pairs, demonstrates limitations in text retrieval tasks. Similarly, certain text representation models specifically designed for dense retrieval tasks do not exhibit robust performance in sentence textual similarity tasks. Recently, there has been a shift in research focus towards developing more comprehensive models for text representation leveraging large quantities of unlabeled web data through unsupervised contrastive pre-training, coupled with task-specific data, prompts, or instructions to mitigate task conflicts during fine-tuning~\citep{ni-etal-2022-sentence, ni-etal-2022-large, neelakantan2022text, wang2022text, INSTRUCTOR}. Additionally, the introduction of benchmarks, such as the Massive Text Embedding Benchmark (MTEB)~\cite{muennighoff-etal-2023-mteb}, has established a robust basis for assessing the universality of text representation models. However, a significant limitation in  existing research is the reliance on in-house data for pre-training, creating a bottleneck in the utilization of pre-trained model weights or APIs. Furthermore, the formulation of prompts specifically tailored for each task requires extra human effort during implementation~\cite{INSTRUCTOR}.

This work presents a straightforward approach to construct a general text embedding (GTE) model solely using contrastive learning on open-source data, as illustrated in Figure \ref{fig:training}. Specifically, we first gather a large-scale dataset comprising unsupervised text pairs extracted from various data sources for contrastive pre-training. Surprisingly, our model, pre-trained on this dataset, exhibits remarkable performance, surpassing BM25 and E5 model~\cite{wang2022text} in zero-shot text retrieval tasks and surpassing many supervised models in the MTEB benchmark. To further enhance the quality of the learned text representations, we obtain high-quality text pairs with human labels from multiple sources for contrastive fine-tuning. After supervised fine-tuning, our $110$M BERT-based~\cite{devlin-etal-2019-bert} model already outperforms the current commercial embedding API of OpenAI and ranks highly in the MTEB benchmark. Furthermore, since our model is trained using code data as well, we evaluate its code search capabilities on the CodeSearchNet benchmark, which encompasses six programming languages. Notably, even without language-specific fine-tuning on each subset, our model significantly outperforms state-of-the-art code retrievers of similar size that have been fine-tuned for each programming language.

In the rest of this paper, we provide a detailed account of the data sources and training configurations employed. Subsequently, we present the evaluation results on widely recognized text embedding benchmarks and compare them with the performance of previous state-of-the-art baselines that were specifically optimized for each individual task. Our model consistently demonstrates superior performance or, at the very least, comparable results to those achieved by larger models, owing to its incorporation of a more diverse mixture of training datasets. We aspire for our model to serve as a robust baseline for the research community investigating text and code embedding.

%% file: sections/rw.tex
\section{Related Work}
Text embeddings serve as low-dimensional vector representations for texts of varying lengths and are essential in numerous natural language processing (NLP) tasks. In contrast to high-dimensional and sparse representations such as TF-IDF, dense text embeddings possess the capacity to address the lexical mismatch problem and enhance the efficiency of text retrieval and matching. 

Pre-trained language models, exemplified by BERT~\cite{devlin-etal-2019-bert} and GPT~\cite{gpt}, have demonstrated remarkable success across various NLP tasks. Nonetheless, extracting a high-quality sentence embedding from pre-trained language models poses a significant challenge due to the presence of anisotropic embedding spaces resulting from the masked language modeling objective. To address this issue, subsequent studies have proposed different approaches, including supervised fine-tuning~\citep{reimers-gurevych-2019-sentence}, normalizing flow~\citep{li-etal-2020-sentence}, normalizing flow~\citep{li-etal-2020-sentence},  whitening~\citep{su2021whitening}, or unsupervised contrastive learning~\citep{gao-etal-2021-simcse}. These investigations primarily concentrate on enhancing performance in semantic textual similarity tasks, wherein two sentences exhibit similar formats.

Another line of research focuses on the text retrieval problem, where the query and document typically exhibit an asymmetric relationship. In this context, the dual-encoder architecture necessitates training with both positive and negative pairs. \citet{lee-etal-2019-latent} propose the Inverse Close Task (ICT) as a self-supervised pre-training approach for generating a dense retriever. The ICT method involves cropping a random sentence from a passage to construct pseudo query-document pairs. Additionally, \citet{Chang2020Pre-training} leverage the link structure within Wikipedia to introduce further supervision signals in the pre-training data. In a similar vein, REALM~\citep{pmlr-v119-guu20a} proposes a joint training approach, wherein a dense retriever and a language model are trained concurrently. The learning signal for the language model is derived from masked language modeling, with backpropagation incorporated through the retrieval step. Recent advancements, such as Contriever~\citep{izacard2022unsupervised} and coCondenser~\citep{gao-callan-2022-unsupervised}, have demonstrated that constructing positive pairs through random passage cropping yields superior results compared to the ICT task. Building upon the ideas presented in ~\citep{Chang2020Pre-training}, some researchers have also put forth methods for constructing higher-quality positive pairs using the web link topology for retriever pre-training~\citep{zhou-etal-2022-hyperlink}, a technique that proves effective in zero-shot scenarios. Furthermore, in the field of dense retrieval, significant research is dedicated to enhancing the text representation capabilities of pre-trained language models through the design of auxiliary pre-training tasks~\cite{Gao2021CondenserAP,Xiao2022RetroMAEPR,gao-callan-2022-unsupervised,Wang2022SimLMPW,rom,li2023cdmae}.

The previous two lines of research can be generalized as learning a vector representation for a piece of text and distinguished by the type of downstream tasks. Recently, several studies have explored the construction of unified text representation models through large-scale contrastive learning and prompt-based learning~\cite{neelakantan2022text,wang2022text,INSTRUCTOR}. Additionally, some research efforts have focused on constructing evaluation datasets to better assess the stability of text representation models across different tasks and domains. BEIR (Benchmarking IR)~\cite{beir} collects a substantial number of retrieval tasks from various domains to evaluate the robustness of dense retriever models in zero-shot scenarios. Meanwhile, MTEB (Massive Text Embedding Benchmark)~\citep{muennighoff-etal-2023-mteb} benchmarks over $56$ datasets spanning seven categories, providing a comprehensive evaluation of text embedding models.

This study aims to develop a general text embedding model through a multi-stage training approach. In the initial stage of unsupervised contrastive learning, we generate weak supervised correlation text pairs using publicly available data from various sources. Unlike previous study~\cite{wang2022text}, we exclusively utilized open-source data and did not employ any filtering or cleaning methods. Pre-training on a large-scale text pairs can effectively improve the domain generalization of text representation models and bridge the gap between the MLM training objective and the contrastive learning objective of representation models, making the language model more suitable for text representation tasks. In the supervised fine-tuning stage, the mixture of training data in our approach is more varied to further enhance the model's versatility. Moreover, our model does not incorporate task-specific prompts, which enhances reproducibility and ease of use.

%% file: sections/approach.tex
\section{Approach}

The training process of our model consists of two stages: unsupervised pre-training and supervised fine-tuning. Both stages employ the learning objective of contrastive learning. Firstly, we will introduce the basic framework of the model. Subsequently, we will discuss the sources and construction methods of the training data in the two stages. Finally, we will present some special optimization strategies used to enhance the model's performance during the training process.

\subsection{Model Architecture}

The backbone of our embedding model is a deep Transformer encoder~\citep{transformer} which can be initialized with pre-trained language models such as BERT~\citep{devlin-etal-2019-bert}.
Our model follows the vanilla dual-encoder architecture with mean pooling on top of the contextualized token representations produced by the language model.

Formally, given a piece of text $x=(x_1,\dots,x_n)$ consisting of $n$ tokens, an embedding model $E$ convert the text into a low-dimensional dense vector $\textbf{x} = E(x) \in R^d$.
To implement $E$, we first employ a language model to get the deep contextualized token representations
\begin{equation}
    \textbf{h} = \text{LM}(x) \in R^{n\times d}.
\end{equation}

Then we apply a lightweight mean pooling across the first dimension to get the text representation,
\begin{equation}
    \textbf{x} = \frac{1}{n}\sum_{i=1}^n \textbf{h}_i \in R^d
\end{equation}

The text representations are learned through the contrastive objective, distinguishing semantic relevant text pairs from irrelevant ones.
Such training procedure requires positive and negative pairs, taking the format of $(q,d^+,d^-)$.
For a query $q$, a relevant document $d^+$, a set of irrelevant documents $\mathcal{D}_-=\{d^-_1,\ldots,d^-_n\}$, one popular contrastive objective is the InfoNCE loss~\citep{infonce},
\begin{equation}
    L_\text{cl} = - \log\frac{e^{s(q,d^+)/\tau}}{e^{s(q,d^+)/\tau}+\sum\limits_{i=1}^{n}e^{s(q,d^-_i)/\tau}},
\end{equation}
where $s(q,d)$ estimates the similarity between two pieces of text $q$ and $d$ via vector distance between $\textbf{q}=E(q)$ and $\textbf{d}=E(d)$.

To acquire text embeddings of superior quality that can be applied across a wide range of scenarios, we compile an extensive text pair dataset from multiple formats and domains. This dataset is then trained using an improved contrastive loss method in a multi-stage fashion. 

\subsection{Unsupervised Pre-training Data}

Weakly supervised text relevance data is readily available in publicly accessible web sources, such as the inherent connection between queries and answers on QA forums. These data can be extensively collected without the need for manual annotation, thereby efficiently aiding in training text representation models. Inspired by previous work~\citep{ni-etal-2022-sentence, ni-etal-2022-large, neelakantan2022text, wang2022text}, our model is initially pre-trained on naturally occurring text pairs extracted from diverse sources. To ensure the versatility of the embedding model, we explore a range of resources for text pair extraction, including web pages (\emph{e.g.}, CommonCrawl, ClueWeb), scientific papers (\emph{e.g.}, arXiv, SemanticScholar), community QA forums (\emph{e.g.}, StackExchange), social media (\emph{e.g.}, Reddit), knowledge bases (\emph{e.g.}, Wikipedia, DBPedia), and code repositories (\emph{e.g.}, StackOverflow, GitHub). Additionally, we harness the presence of hyperlinks in certain datasets to facilitate text pair extraction. Table~\ref{tab:pt_data} demonstrates some examples of text pair format from different sources. Further details regarding the data collection process can be found in Appendix~\ref{sec:appendix}. In total, we utilized $\sim$800M text pairs text pairs for the unsupervised pre-training stage. Simple statistics and data distributions are illustrated in Table~\ref{tab:pretrain}.

\begin{table}[!ht]
    \centering
    \begin{tabular}{lcrr}
    \toprule
    Source & Datasets & Prop. & Size \\
    \midrule
    Web Page & 3 & 18.7\% & 147M \\
    Academic Paper & 5 & 5.7\% & 45M \\
    Hyperlink & 4 & 13.4\% & 106M \\
    Social Media & 2 & 41.5\% & 327M \\
    Knowledge Base & 2 & 4.8\% & 38M \\
    Community QA & 7 & 1.5\% & 12M \\
    News & 5 & 0.4\% & 3M \\
    Code & 2 & 2.5\% & 20M \\
    Others & 3 & 11.6\% & 91M \\
    \midrule
    Total & 33 & 100\% & 788M \\
    \bottomrule
    \end{tabular}
    \caption{Statistics of pre-training data.}
    \label{tab:pretrain}
\end{table}

\input{tables/pt_data}

\subsection{Supervised Fine-tuning Data}

In the supervised fine-tuning stage, we use relatively lower-sized datasets with human annotation of the relevance between two pieces of text and optional hard negatives mined by an extra retriever to form text triples.
To handle both symmetric tasks (\emph{e.g.}, semantic textual similarity) and asymmetric tasks (\emph{e.g.}, passage retrieval), we collect data from a large variety of tasks and domains, including web search (\emph{e.g.}, MS MARCO), open-domain QA (\emph{e.g.}, NQ), NLI (\emph{e.g.}, SNLI), fact verification (\emph{e.g.}, FEVER), paraphrases (\emph{e.g.}, Quora).
We totally used $\sim$3M pairs for fine-tuning, which is a combination of training data used by previous research~\citep{gao-etal-2021-simcse, gao-callan-2022-unsupervised, tart, INSTRUCTOR, li2023cdmae}.
More details can be found in Appendix~\ref{sec:appendix}.

\subsection{Training Details}

\paragraph{Data Sampling}
In the initial stage of unsupervised pre-training, data sources often differ significantly in terms of the number of training instances. To address this imbalance, we employ a multinomial distribution to sample data batches from different data sources, taking into account their respective sizes. Suppose the whole pre-training dataset $D$ consists of $m$ different subsets $\{D_1,\ldots,D_m\}$ and denote the size of each subset as $n_i=|D_i|$, at each training iteration, the probability of sampling data from the $i$-th subset $D_i$ can be represented by:
\begin{equation}
    p_i = \frac{n_i^\alpha}{\sum_{j=1}^{m}n_j^\alpha},
\end{equation}
where we set $\alpha=0.5$ in this work. Furthermore, to prevent the model from solely learning task-specific shortcuts for discrimination, we ensure that all training instances within a batch originate from the same task.

\paragraph{Improved Contrastive Loss}
When using the contrastive objective, people usually reuse in-batch documents as negative candidates to improve training efficiency~\cite{karpukhin-etal-2020-dense}.
This paper uses an improved contrastive learning objective which is bidirectional and enlarges the negative samples with both in-batched queries and documents.
This can be viewd as a combination of loss variants proposed by~\citet{clip, ren-etal-2021-pair, samtone}.

Consider a batch of positive text pair samples $$B=\{(q_1,d_1),(q_2,d_2),...,(q_n,d_n)\},$$
we use an improved contrastive loss which takes the form
\begin{equation}
\label{equ:icl}
    L_{\text{icl}} = -\frac{1}{n} \sum_{i=1}^n \log \frac{e^{s(q_i, d_i)/\tau}}{Z}
\end{equation}
with the partition function being
\begin{equation}
\begin{aligned}
    Z = \sum_{j}e^{s(q_i,d_j)/\tau}+\sum_{j\neq i}e^{s(q_i,q_j)/\tau} \\
    +\sum_{j}e^{s(q_j,d_i)/\tau}+\sum_{j\neq i}e^{s(d_j,d_i)/\tau}
\end{aligned}
\end{equation}
in which the first two terms are used for query to document contrast, where as the last two terms are used for the inverse.
In this work, we use the cosine similarity as the distance metric
\begin{equation}
    s(q,d) = \frac{\textbf{q}\cdot\textbf{d}}{||\textbf{q}||_2\cdot||\textbf{d}||_2}.
\end{equation}
The temperature $\tau$ is fixed to 0.01 in this work.

\paragraph{Training and Evaluation}

\begin{table*}[!ht]
    \centering
    \begin{tabular}{lrcccc}
    \toprule
    Model & Params & LR & GPUs & BS & Base LM \\
    \midrule
    GTE$_\text{small}$ & 30M & $3\times 10^{-4}$ & 2 & 16384 & \texttt{microsoft/MiniLM-L12-H384-uncased} \\
    GTE$_\text{base}$ & 110M & $2\times 10^{-4}$ & 4 & 16384 & \texttt{bert-base-uncased} \\
    GTE$_\text{large}$ & 330M & $5 \times 10^{-5}$ & 8 & 16384 & \texttt{bert-large-uncased} \\
    \bottomrule
    \end{tabular}
    \caption{Pre-training configurations of models of different sizes.}
    \label{tab:config}
\end{table*}

The training of our embedding model consists of two stages.
In the first stage of contrastive pre-training with only in-batch negatives, using a large batch size is crucial to better model performance by reducing the gap between training and inference with more negatives included and providing a better approximation to the underlying learning objective.
To facilitate this, we limit the maximum sequence length to $128$ during pre-training and distribute the use of negatives across all GPUs.
Popular techniques such as automatic mixed precision training~\citep{micikevicius2018mixed} with fp16, deepspeed ZeRO~\citep{zero} stage 1 and gradient checkpointing~\citep{chen2016training} are also jointly used to reduce memory cost and scale up batch size to over ten thousands.
We run the pre-training for $50,000$ steps, which roughly corresponds to one epoch on the whole pre-training data.
We only tuned the learning rate to ensure the convergence of larger models.
we employ the AdamW optimizer with linear learning rate decay and a warm-up period during the initial $5\%$ of training steps. We conducted experiments on three distinct model scales: small, base, and large. These models were initialized using the small-sized MiniLM~\cite{minilm} model and the base and large models of the BERT~\cite{devlin-etal-2019-bert} model. Further details can be found in Table~\ref{tab:config}.

In the second stage of contrastive fine-tuning with supervised data and hard negatives, a large batch size is unnecessary since hard negatives can already provide a reliable gradient estimation of the learning objective~\citep{xiong2021approximate, li2023cdmae}.
Therefore, a global batch size of 128 and a train group size of 16 are utilized, with one positive example and the remaining being either hard negatives or random negatives. 
Instead we increase the max sequence length to 512 to better handle texts with longer lengths.
The learning rate is decreased by a factor of ten during fine-tuning.
The model is fine-tuned on the collected dataset for a single epoch. In-batch texts are also incorporated as negative candidates using the enhanced contrastive loss described in Equation~\ref{equ:icl}.

After training, we directly take the last checkpoint for evaluation.
We run model training on up to 8 NVIDIA A100 GPUs with 80GB memory and model evaluation on up to 8 NVIDIA Tesla V100 GPUs with 32GB memory.
Models are trained with mixed precision using fp16 and evaluated with half precision fp16 as well.

%% file: tables/pt_data.tex
\begin{table*}[ht]
\centering
\resizebox{\textwidth}{!}{
\begin{tabular}{llll}
\hline
Task Type & Text Pair Format & Query & Doc \\
\hline
Web Page & \begin{tabular}[c]{@{}l@{}}(title, body)\end{tabular} & \begin{tabular}[c]{@{}l@{}}Providence Real Estate | Providence Homes for Sale\end{tabular} & \begin{tabular}[c]{@{}l@{}}Founded by Roger Williams in 1636, Providence is \\ recognized as one of the country's oldest cities\ldots\end{tabular} \\
\hline
Academic Paper &  \begin{tabular}[c]{@{}l@{}}(title, abstract)\end{tabular} & \begin{tabular}[c]{@{}l@{}}Polymer Quantum Mechanics and its Continuum Limit\end{tabular} & \begin{tabular}[c]{@{}l@{}}A rather non-standard quantum representation of the 
\\ canonical commutation relations of quantum mechanics\ldots\end{tabular}   \\ \hline
Hyperlink &  \begin{tabular}[c]{@{}l@{}}(citation, reference)\end{tabular} & \begin{tabular}[c]{@{}l@{}}After the championship in 1996, the PGA of America \\ raised its stake to 50\% and announced that \ldots\end{tabular} &  \begin{tabular}[c]{@{}l@{}}Pebble Beach Golf Links The largest margin of victory \\ ever in a major championship, surpassing the 13-shot \ldots\end{tabular} \\ \hline
Social Media & (post, comment) & \begin{tabular}[c]{@{}l@{}}Pretty sure any team with Lebron James will be a playoff \\ contender. Considering UNC would be in the East\ldots \end{tabular} &  \begin{tabular}[c]{@{}l@{}}I was being sarcastic and making fun of the East, but \\ honestly I was really in deep thought about this \ldots \end{tabular} \\
\hline
Knowledge Base & \begin{tabular}[c]{@{}l@{}}(entity, description)\end{tabular}  & \begin{tabular}[c]{@{}l@{}}Animation \end{tabular}   & \begin{tabular}[c]{@{}l@{}}Animation is the process of creating the illusion of motion \\ and shape change by means of the rapid display of \ldots \end{tabular} \\
\hline
Community QA & (question, answer) & \begin{tabular}[c]{@{}l@{}}How the human species evolved? \end{tabular} & \begin{tabular}[c]{@{}l@{}}A tough question as it overlaps science and theology. Since \\ you asked ``how the human species evolved?'' I'll assume \ldots \end{tabular} \\
\hline
News  & \begin{tabular}[c]{@{}l@{}}(summary, content)\end{tabular}  & \begin{tabular}[c]{@{}l@{}}Nepalese Opposition Welcomes Return of Parliament \end{tabular}   &  \begin{tabular}[c]{@{}l@{}}Nepal's opposition alliance formally calls off weeks of \\ pro-democracy protests after King Gyenandra reinstates \ldots \end{tabular} \\ \hline
Code & \begin{tabular}[c]{@{}l@{}}(text, code)\end{tabular} & \begin{tabular}[c]{@{}l@{}}SetMaxRecords sets the MaxRecords field's value. \end{tabular}   & \begin{tabular}[c]{@{}l@{}}func (s *DescribeSnapshotCopyGrantsInput) SetMaxRecords \\(v int64) *DescribeSnapshotCopyGrantsInput \{ s.MaxRecords \end{tabular}  \\ \hline
\end{tabular}
}
\caption{Examples of mined (query, document) pairs in the pre-training data.}
\label{tab:pt_data}
\end{table*}

%% file: sections/results.tex
\section{Experiments}

In this section, we provide an extensive evaluation of our embedding model, comparing to state-of-the-art models for each task.
Note that an apple-to-apple comparison is hardly possible since different models used different in-house data for pre-training and the base language models vary a lot.
We mainly use the number of model parameters as a criterion for performance comparison since it is closely related to the inference speed.

\subsection{Zero-shot Text Classification}
\input{tables/zeroshot_sst2}
One method to assess the quality of learned representation is through zero-shot classification.~\citep{clip, neelakantan2022text, wang2022text}.
We recast text classification into an embedding-based similarity matching problem.
In this setting, inputs texts are converted into embeddings directly and labels are verbalized to corresponding text to get label embeddings.
Distances between input embeddings and label embeddings are measured by their inner product and label with the most close embedding distance to the input text is regarded as the classification result.
An example is SST-2 binary sentiment classification task.
We consider two types of label verbalizers for evaluation.
The vanilla version uses the sentiment word `positive' or `negative' to denote the corresponding labels.
Prompted version uses fuzzy prompt template, such as `this is an example of positive/negative movie review'.

Zero-shot text classification accuracy on SST-2 is shown in Table~\ref{tab:zeroshot_sst2}.
In the vanilla setting, our 110M model already matches the performance of prompted E5$_\text{large}$ with 330M parameters.
Using prompting strategy further improves results significantly and closes the gap with large models.
Even without explicit prompt or instruction during training, our model can somewhat understand the label context better when formatted as a natural language text.

\subsection{Unsupervised Text Retrieval}
Text retrieval requires retrieving most relevant documents from a large-scale candidate sets.
We use BEIR~\cite{beir} as our evaluation benchmark for zero-shot unsupervised text retrieval.
BEIR is a heterogeneous information retrieval benchmark which contains retrieval tasks of different formats and from different domains.
We use the open available 15 datasets for evaluation.

\begin{figure*}[!ht]
    \centering
    \includegraphics[width=\textwidth]{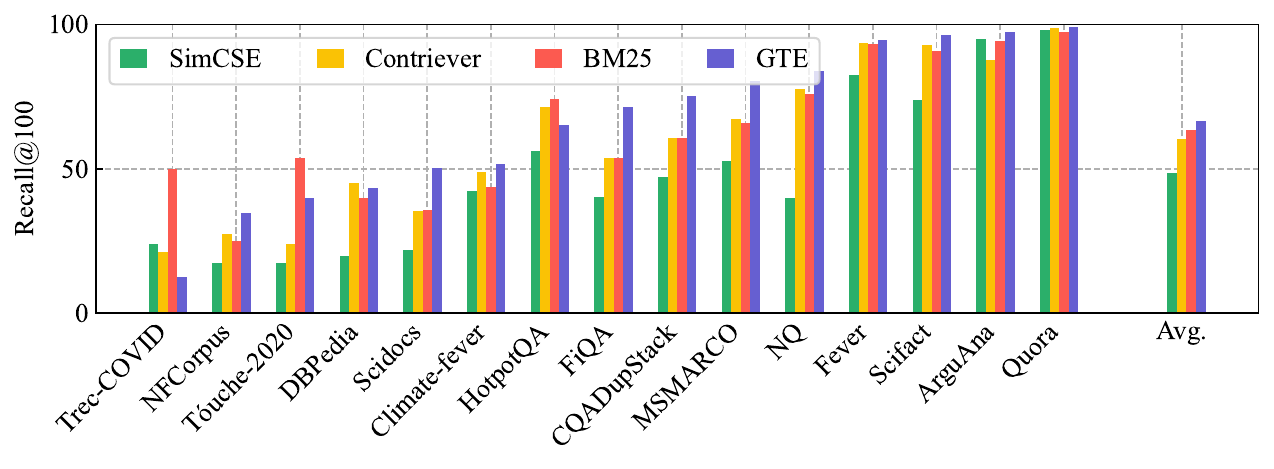}
    \caption{
    Recall@100 of unsupervised text retrieval methods on BEIR benchmark~\citep{beir}.
    We compare our model GTE$_\text{base}$ (based on BERT$_\text{base}$) without using any annotated data to SimCSE~\citep{gao-etal-2021-simcse} (based on RoBERTa$_\text{large}$), Contriever~\citep{izacard2022unsupervised} (based on BERT$_\text{base}$) and BM25.
    Baseline results are borrowed from the Contriever paper~\citep{izacard2022unsupervised} with dot product being the similarity function.
    }
    \label{fig:beir_unsup_recall}
\end{figure*}

\input{tables/beir_unsup}

We compare our unsupervised pre-trained checkpoint to recent unsupervised dense retrievers such as Contriever~\cite{izacard2022unsupervised} and E5~\cite{wang2022text}.
According to Table~\ref{tab:beir_unsup_results}, we find that our base size model significantly outperforms the models with comparable size, like SimCSE, Contriever and E5.
Our base model is comparable to E5$_\text{large}$ without using human supervision.

\subsection{Massive Text Embedding Benchmark}

Massive Text Embedding Benchmark (MTEB) is a comprehensive semi-supervised benchmark that incorporates a limited amount of supervision data for evaluation. In this paper, we evaluate the English subsets which encompasses 56 English datasets across seven distinct tasks, including text classification (Class.), text clustering (Clust.), pairwise classification (Pair.), text reranking (Rerank.), text retreival (Retr.), semantic textual similarity (STS) and summarization (Summ.). The evaluation metrics employed in MTEB are accuracy, v-measure, average precision, MAP, nDCG@10, and Spearman coefficients, respectively. For further details on the tasks covered in the MTEB benchmark, please refer to the Appendix~\ref{sec:mteb}.

\input{tables/mteb}

Two settings are considered for comparison: the unsupervised setting and the supervised setting. In the unsupervised setting, models are trained using unlabeled data, while supervised models are fine-tuned using high-quality datasets with human labels. The results of strong baseline models are presented in Table~\ref{tab:mteb}.

In the unsupervised setting, our model outperforms the previous best model, E5, by a significant margin across all considered tasks, without the use of task-specific prompts. This improvement can be attributed to the inclusion of more training data formats and various sources of self-supervision signals. Furthermore, it is worth noting that our unsupervised pre-trained model narrows the gap even further with larger supervised baselines, such as GTR and Sentence-T5. In the supervised setting, our model surpasses OpenAI results by a large margin despite using a modest model size. GTE$_\text{small}$ is comparable to E5$_\text{large}$ while being 10$\times$ smaller. GTE$_\text{large}$ establishes new state-of-the-art performance on the MTEB benchmark, outperforming the multi-task instruction-finetuned embedding model, InstructOR$_\text{large}$, by 1.5 points on average.

\subsection{Code Search}

\input{tables/codesearch_hard}

Programming languages can be regarded as a distinct form of text. To assess the effectiveness of our approach in code search, we conduct a comparative analysis with other code-based language models, such as CodeBERT~\citep{guo2021graphcodebert} and GraphCodeBERT~\citep{guo2021graphcodebert}. We also compare our approach with a more recent code language model called UniXcoder~\citep{guo-etal-2022-unixcoder}, which aims to integrate various pre-training tasks into a unified model. CodeRetriever~\citep{li-etal-2022-coderetriever} is initialized from GraphCodeBERT and pre-trained on large-scale multi-modal code-text pairs mined and cleaned by heuristics. It is important to note that while the baseline models are individually trained and evaluated for each programming language, our model is directly evaluated across all the languages.

In line with recent work~\cite{guo2021graphcodebert, guo-etal-2022-unixcoder, li-etal-2022-coderetriever}, we mainly evaluate on the challenging settings where the code corpus include all codes from dev and test set instead of 1k randomly sampled code.\footnote{
Evaluation on the original settings can be found in Appendix~\ref{app:csn}.
}
The results are presented in Table~\ref{table:code-full}. Surprisingly, our model surpasses models that are pre-trained on code and then fine-tuned for each programming language separately. This finding demonstrates that, by scaling the amount of data and computational resources, the language model can acquire high-quality code representations directly from sequences of code tokens, without the need for incorporating human knowledge about the structural information of code~\citep{guo2021graphcodebert}. We observe a significant improvement in Python, likely due to its resemblance to natural language. Our model, pre-trained on an extensive text pairs spanning various domains, demonstrates effective cross-task knowledge transfer from text retrieval to code retrieval.

%% file: tables/zeroshot_sst2.tex
\begin{table}[!ht]
    \centering
    \begin{tabular}{lrcc}
    \toprule
    Model & Params & Prompting & Accuracy \\ \midrule
    E5$_\text{base}$ & 110M & \cmark & 81.3 \\ 
    E5$_\text{large}$ & 330M & \cmark & 85.3 \\
    \texttt{cpt-text} & 6B & & 88.1 \\
    \texttt{cpt-text} & 6B & \cmark & 89.1 \\
    \midrule
    GTE$_\text{base}$ & 110M & & 85.1 \\
    GTE$_\text{base}$ & 110M & \cmark & 87.2 \\
    \bottomrule
    \end{tabular}
    \caption{Zero shot text classification performance on SST-2. All compared models are the fine-tuned ones.}
    \label{tab:zeroshot_sst2}
\end{table}

%% file: tables/beir_unsup.tex
\begin{table*}[!ht]
\centering
\setlength{\tabcolsep}{2pt}
\begin{tabular}{lccccccccccc}
\toprule
Dataset & BM25 & SimCSE & Contriever & CPT-S & E5$_\text{small}$ & E5$_{\text{base}}$ & E5$_{\text{large}}$ & GTE$_\text{small}$ & GTE$_\text{base}$ & GTE$_\text{large}$ \\
\midrule
MS MARCO & 22.8 & 9.4 & 20.6 & 19.9 & 25.4 & 26.0 & 26.2 & 31.3 & 31.8 & 31.7 \\
Trec-Covid & 65.6 & 26.2 & 27.4 & 52.9 & 52.0 & 61.0 & 61.8 & 61.8 & 64.0 & 64.8 \\
NFCorpus & 32.5 & 9.9 & 31.7 & 32.0 &  29.3  & 35.8 & 33.7 & 34.9 & 36.2 & 38.1 \\
NQ & 32.9 & 11.7 & 25.4 & - &  37.3  & 39.0 & 41.7 & 32.0 & 35.3 & 34.5 \\
HotpotQA & 60.3 & 19.8 & 48.1 & 51.5 &  46.0  & 52.4 & 52.2 & 49.3 & 50.8 & 49.2 \\
FiQA & 23.6 & 9.8 & 24.5 & 34.1 &  38.3  & 40.0 & 43.2 & 37.0 & 36.9 & 40.6 \\
ArguAna & 31.5 & 38.3 & 37.9 & 38.7 &  42.5  & 42.2 & 44.4 & 41.6 & 41.0 & 41.3 \\
Touche-2020 & 36.7 & 8.9 & 19.3 & 21.0 & 19.9  & 16.9 & 19.8 & 17.7 & 18.2 & 18.5 \\
CQADupStack & 29.9 & 13.2 & 28.4 & - & 35.0  & 35.4 & 38.9 & 38.1 & 39.9 & 39.8 \\
Quora & 78.9 & 78.0 & 83.5 & 68.1 &  85.8  & 85.7 & 86.1 & 86.1 & 85.0 & 84.8 \\
DBPedia & 31.3 & 15.0 & 29.2 & 27.2 & 34.5  & 35.4 & 37.1 & 33.5 & 33.2 & 33.6 \\
Scidocs & 15.8 & 5.5 & 14.9 & - &  19.9  & 21.1 & 21.8 & 21.5 & 22.5 & 22.7 \\
Fever & 75.3 & 21.1 & 68.2 & 57.1 &  62.5  & 63.4 & 68.6 & 71.3 & 72.7 & 70.5 \\
Climate-Fever & 21.3 & 11.8 & 15.5 & 15.8 & 14.5 & 15.4 & 15.7 & 21.4 & 21.0 & 25.4 \\
Scifact & 66.5 & 25.7 & 64.9 & 65.4 & 68.5  & 73.7 & 72.3 & 72.7 & 74.1 & 74.1 \\
\midrule
Average & 41.7 & 20.3 & 36.0 & - & 40.8  & 42.9 & 44.2 & 43.4 & 44.2 & 44.6 \\
\bottomrule
\end{tabular}
\caption{nDCG@$10$ of different unsupervised methods on the BEIR benchmark~\citep{beir}.
SimCSE is based on BERT$_\text{base}$ backbone.
CPT-S~\citep{neelakantan2022text} is of similar size to BERT$_\text{large}$.
Baseline results are borrowed from E5 paper~\citep{wang2022text}.
Note that Contriever uses dot product as the similarity metric while other models uses cosine similarity.
}
\label{tab:beir_unsup_results}
\end{table*}

%% file: tables/mteb.tex
\begin{table*}[!ht]
\centering
\begin{tabular}{lrcccccccc}
\toprule
 & Params & Class. & Clust. & Pair. & Rerank & Retr. & STS & Summ. & Avg \\
\multicolumn{1}{l}{\# of datasets $\rightarrow$} & & 12     & 11     & 3          & 4      & 15    & 10  & 1    & 56  \\ \hline
\multicolumn{9}{l}{\emph{Unsupervised models}}    \\ \hline
\multicolumn{1}{l}{Glove}   & 120M &  57.3   &  27.7 & 70.9  & 43.3  & 21.6 & 61.9 & 28.9  & 42.0 \\
\multicolumn{1}{l}{BERT}    & 110M &   61.7  & 30.1  &  56.3 & 43.4  &  10.6  &  54.4   &  29.8  &  38.3  \\
\multicolumn{1}{l}{SimCSE}   & 110M & 62.5  &  29.0  &  70.3   &  46.5  & 20.3  & 74.3 & 31.2  & 45.5 \\
\multicolumn{1}{l}{E5$_\text{small}$}   & 30M &   67.0  & 41.7  &  78.2   &  53.1  &  40.8  & 68.8  & 25.2  &  54.2  \\
\multicolumn{1}{l}{E5$_\text{base}$}   & 110M &   67.9  & 43.4  &  79.2   &  53.5  &  42.9  & 69.5  & 24.3  &  55.5  \\
\multicolumn{1}{l}{E5$_\text{large}$}  & 330M & 69.0  &  44.3  &  80.3   &  54.4  & 44.2  & 69.9 & 24.8  & 56.4 \\ 
\hline
\multicolumn{1}{l}{GTE$_\text{small}$} & 30M & 71.0 & 44.9 & 82.4 & 57.5 & 43.4 & 77.2 & 30.4 & 58.5 \\
\multicolumn{1}{l}{GTE$_\text{base}$} & 110M & 71.5 & 46.0 & 83.3 & 58.4 & 44.2 & 76.5 & 29.5 & 59.0 \\
\multicolumn{1}{l}{GTE$_\text{large}$} & 330M & 71.8 & 46.4 & 83.3 & 58.8 & 44.6 & 76.3 & 30.1 & 59.3 \\
\hline
\multicolumn{9}{l}{\emph{Supervised models}}  \\
\hline
\multicolumn{1}{l}{SimCSE}  & 110M & 67.3  &  33.4  &  73.7   &  47.5  & 21.8  &  79.1    &  23.3    &  48.7 \\
\multicolumn{1}{l}{Contriever}   & 110M &  66.7  &  41.1  &  82.5  & 53.1  & 41.9   &  76.5   &  30.4    &  56.0   \\
\multicolumn{1}{l}{GTR$_\text{large}$}   & 330M & 67.1    &  41.6   &   85.3   &  55.4  & 47.4  & 78.2 & 29.5  &  58.3   \\
\multicolumn{1}{l}{Sentence-T5$_\text{large}$}   & 330M & 72.3  &  41.7   &  85.0   &  54.0  & 36.7  &  81.8 &  29.6  & 57.1 \\
\multicolumn{1}{l}{E5$_\text{small}$}     & 30M & 71.7 &  39.5  &  85.1  & 54.5  & 46.0  & 80.9  & 31.4 & 58.9 \\
\multicolumn{1}{l}{E5$_\text{base}$}     & 110M &  72.6 &  42.1  &  85.1  & 55.7  & 48.7 & 81.0  & 31.0 & 60.4 \\
\multicolumn{1}{l}{E5$_\text{large}$}     & 330M & 73.1 & 43.3  &  85.9  & 56.5 & 50.0  & 82.1 & 31.0 & 61.4  \\
\multicolumn{1}{l}{InstructOR$_\text{base}$}     & 110M & 72.6 & 42.1 & 85.1 & 55.7 & 48.8 & 81.0 & 31.0 & 60.4 \\
\multicolumn{1}{l}{InstructOR$_\text{large}$}     & 330M & 73.9 & 45.3 & 85.9 & 57.5 & 47.6 & 83.2 & 31.8 & 61.6 \\
\multicolumn{1}{l}{OpenAI$_\text{ada-001}$}   & n.a. & 70.4 & 37.5 & 76.9 & 49.0 & 18.4 & 78.6 & 26.9 & 49.5 \\
\multicolumn{1}{l}{OpenAI$_\text{ada-002}$}   & n.a. & 70.9 & 45.9 & 84.9 & 56.3 & 49.3 & 81.0 & 30.8 & 61.0 \\
\hline
\multicolumn{1}{l}{GTE$_\text{small}$} & 30M & 72.3 & 44.9 & 83.5 & 57.7 & 49.5 & 82.1 & 30.4 & 61.4 \\
\multicolumn{1}{l}{GTE$_\text{base}$}     & 110M & 73.0 & 46.1 & 84.3 & 58.6 & 51.2 & 82.3 & 30.7 & 62.4 \\
\multicolumn{1}{l}{GTE$_\text{large}$}     & 330M & 73.3 & 46.8 & 85.0 & 59.1 & 52.2 & 83.4 & 31.7 & 63.1 \\
\hline
\multicolumn{9}{l}{\emph{Larger models}}        \\ \hline
\multicolumn{1}{l}{InstructOR$_\text{xl}$}   & 1.5B & 73.1 & 44.7 & 86.6 & 57.3 & 49.3 & 83.1 & 32.3 & 61.8 \\
\multicolumn{1}{l}{GTR$_\text{xxl}$}   & 4.5B & 67.4  &  42.4   & 86.1   &  56.7   &  48.5  &  78.4   & 30.6  &  59.0 \\
\multicolumn{1}{l}{Sentence-T5$_\text{xxl}$}   & 4.5B & 73.4   &  43.7  &  85.1    &  56.4  & 42.2  & 82.6  & 30.1  &  59.5 \\
\bottomrule
\end{tabular}
\caption{
Results on the MTEB~\citep{muennighoff-etal-2023-mteb} (56 datasets in English subset).
Compared models include SimCSE~\citep{gao-etal-2021-simcse}, Sentence-T5~\citep{ni-etal-2022-sentence}, GTR~\citep{ni-etal-2022-large}, Contriever~\citep{izacard2022unsupervised}, OpenAI text embedding API~\citep{neelakantan2022text},  E5~\citep{wang2022text} and InstructOR~\citep{INSTRUCTOR}.
Exact parameter amount of OpenAI ada model is not available, but is suspected to be $\sim$300M, comparable to the BERT large size model.
}
\label{tab:mteb}
\end{table*}

%% file: tables/codesearch_hard.tex
\begin{table*}[!ht]
\setlength{\tabcolsep}{5pt}
\centering
\begin{tabular}{lcccccccc}
\toprule
Model & Params & Ruby & JS & Go & Python & Java & PHP & Avg. \\
\midrule
CodeBERT & 110M$\times$6 & 67.9 & 62.0 & 88.2 & 67.2 & 67.6 & 62.8 & 69.3 \\
GraphCodeBERT & 110M$\times$6 & 70.3 & 64.4 & 89.7 & 69.2 & 69.1 & 64.9 & 71.3 \\
UniXcoder & 110M$\times$6 & 74.0 & 68.4 & 91.5 & 72.0 & 72.6 & 67.6 & 74.4 \\
CodeRetriever & 110M$\times$6 & 77.1 & 71.9 & 92.4 & 75.8 & 76.5 & 70.8 & 77.4 \\
\midrule
GTE$_\text{base}$ & 110M & 76.1 & 73.6 & 88.1 & 95.9 & 80.1 & 85.3 & 83.2 \\
\bottomrule
\end{tabular}
\caption{
Results on CodeSearchNet.
Comparison on code search across 6 programming languages~\cite{codesearchnet} with CodeBERT~\cite{feng-etal-2020-codebert}, GraphCodeBERT~\cite{guo2021graphcodebert}, UniXcoder~\cite{guo-etal-2022-unixcoder} and CodeRetriever~\cite{li-etal-2022-coderetriever}.
This setting requires finding the corresponding code candidates from all candidates from dev and test set.
}
\label{table:code-full}
\end{table*}

%% file: sections/analysis.tex
\section{Analysis}

In this section, we analyze the crucial factors influencing model performance and present a series of ablation experiments. Unless otherwise stated, the experiments are performed using a BERT-base scale model with 110M parameters. The training steps and epochs remain consistent across all ablation experiments.

\subsection{Impact of Scaling}
We investigate the impact of scaling the number of data sources, batch size, and model parameters on the quality of learned text embeddings. The evaluation is conducted on the MTEB benchmark.

\begin{figure*}[!ht]
    \centering
\begin{subfigure}{0.32\textwidth}
    \begin{tikzpicture}
    \begin{axis}[
      xlabel={Level},         
      xtick={1, 2, 3},        
      xticklabels={+, ++, +++}, 
      xtick style={draw=none}, ytick style={draw=none},
      ymin=55, ymax=65,
      width=\textwidth,
      legend cell align = {left},
      legend pos = south east,
    ]
    \addplot[
      blue,
      mark=*,
    ] coordinates {
      (1, 56.8)    
      (2, 58.1)    
      (3, 59.0)   
    };
    \addlegendentry{\tiny PT}
    \addplot[
      green,
      mark=*,
    ] coordinates {
      (1, 61.52)    
      (2, 62.06)    
      (3, 62.39)   
    };
    \addlegendentry{\tiny FT}
    \end{axis}
  \end{tikzpicture}
    \caption{Number of training datasets.}
    \label{fig:multitask}
\end{subfigure}
\hfill
\begin{subfigure}{0.32\textwidth}
    \begin{tikzpicture}
\begin{axis}[
xmode=log,
ybar,
log basis x=2,
ymin = 55, ymax = 65,
xlabel={$B$},
width = \textwidth,
xtick=data, 
xtick style={draw=none},
ytick style={draw=none},
bar width=0.45cm,
]
\addplot [pattern=north west lines, pattern color=orange!80] coordinates {
    (2048, 56.78)
    (4096, 57.27)
    (8192, 58.99)
    (16384, 59.03)
};
\addlegendentry{\tiny PT}
\label{bs}
\end{axis}
\end{tikzpicture}
    \caption{Batch size.}
    \label{fig:bs}
\end{subfigure}
\hfill
\begin{subfigure}{0.32\textwidth}
\begin{tikzpicture}
    \begin{axis}[
    xmode=log,
    log basis x=10,
    ymin = 55, ymax = 65,
    xlabel={$N$},
    xticklabels={, 30M, 110M, 330M},
    xtick style={draw=none}, ytick style={draw=none},
    width = \textwidth,
    legend cell align = {left},
    legend pos = south east,
]
\addplot [blue, mark=square] coordinates {
    (30000000, 58.5)
    (110000000, 59.0)
    (330000000, 59.3)
};
\label{num_params}
\addlegendentry{\tiny PT}
\addplot [green, mark=square] coordinates {
    (30000000, 61.4)
    (110000000, 62.4)
    (330000000, 63.1)
};
\label{num_params}
\addlegendentry{\tiny FT}
\end{axis}
    \end{tikzpicture}
    \caption{Number of model parameters.}
    \label{fig:parameters}
\end{subfigure}
\caption{Scaling analysis of different factors during contrastive pre-training and fine-tuning. Model performance is mesured by average performance on MTEB.}
\label{fig:scaling}
\end{figure*}
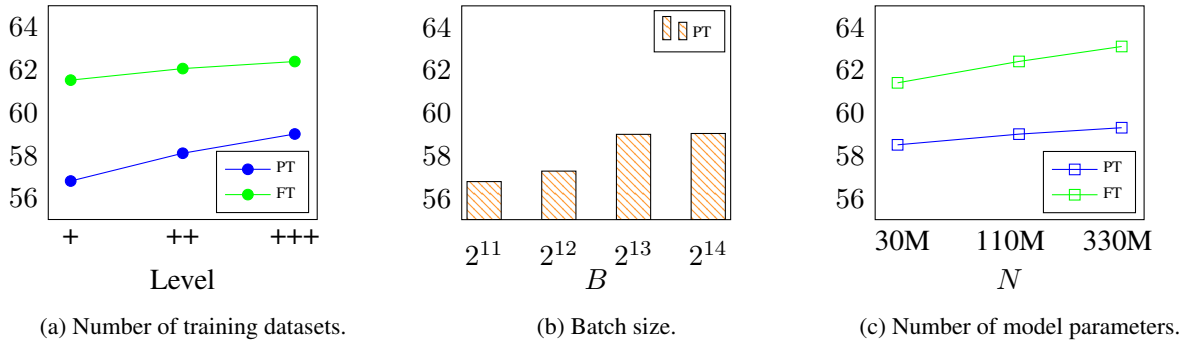

\paragraph{Number of Training Datasets}
First, we conducted an ablation study on the number of training datasets used in pre-training. Model training was carried out by randomly sampling a subset from all available datasets. In the pre-training stage, the first group consisted of only the five largest datasets, ranked by size. The second group included an additional 10 randomly sampled datasets, resulting in a mixture of 15 datasets. The third group utilized all 33 datasets in the pre-training process. For fine-tuning, we initially started with the three datasets used in E5~\citep{wang2022text} fine-tuning and gradually incorporated datasets from MEDI~\citep{INSTRUCTOR} and BERRI~\citep{tart} to investigate the potential benefits. The results presented in Figure~\ref{fig:multitask} demonstrate that the inclusion of more diverse data sources consistently enhances model performance during both the pre-training and fine-tuning stages.

\paragraph{Pre-training Batch Size}
We gradually increase the batch size by a factor of 2 while keeping the training steps fixed to study the influence of batch size used in embedding model pre-training.
According to Figure~\ref{fig:bs}, model performance saturates at around a batch size of ten thousands.
No performance gain is observed when further scaling up batch size.

\paragraph{Number of Model Parameters}
We investigate the scaling behavior by training language models of various sizes, including 30M, 110M, and 330M, which correspond to the small, base, and large scales of the BERT model. Figure~\ref{fig:parameters} illustrates the performance of the pre-trained and fine-tuned models. It can be observed that as the model size grows exponentially, the model performance also improves linearly.

\subsection{Training Behavior}

\begin{figure}
    \centering
    \includegraphics[scale=0.4]{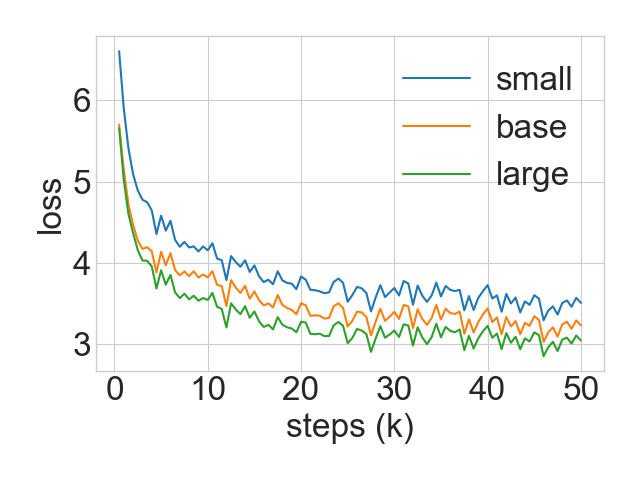}
    \caption{Loss during contrastive pre-training for models of different size.}
    \label{fig:loss}
\end{figure}

We plot the training loss of different sized models during contrastive pre-training in Figure~\ref{fig:loss}.
Larger models have better ablity at learning to distinguish positive pairs from negative ones.
The training loss experiences minor fluctuations consistently across all model scales, which suggests variations in the quality and difficulty of data per batch.\footnote{We use a fixed random seed for data sampling during model training, ensuring that each model encounters the data batches in the same order.}

We also evaluate model performance at different training steps.
It's shown that model performance saturates at 20k steps roughly corresponds to training convergence.

\begin{table}[!ht]
    \centering
    \begin{tabular}{ccccccc}
    \toprule
    Steps & 10k & 20k & 30k & 40k & 50k \\
    \midrule
    MTEB & 56.4 & 59.0 & 57.8 & 57.7 & 59.0 \\
    \bottomrule
    \end{tabular}
    \caption{Model performance at different training steps during unsupervised contrastive pre-training.}
    \label{tab:pt_steps}
\end{table}

\subsection{Influence of Different Training Stages}
To examine the efficacy of multi-stage contrastive learning, we conducted an analysis on the training strategies. We compared three settings: a) solely pre-training on unsupervised text pairs extracted from diverse sources; b) solely fine-tuning on supervised datasets; c) contrastive pre-training followed by fine-tuning. All models were initialized from the original BERT base model.

\begin{table}[!ht]
    \centering
    \begin{tabular}{cccc}
    \toprule
    Setting & PT & FT & Full \\
    \midrule
    MTEB & 59.0 & 57.8 & 62.4 \\
    \bottomrule
    \end{tabular}
    \caption{Model performance at different training stages. PT denotes run only unsupervised pre-training. FT only use supervised data for model trainining. Full apply two stages in a sequential manner.}
    \label{tab:analysis_trainining_stage}
\end{table}

It can be observed from Table~\ref{tab:analysis_trainining_stage} that relying solely on supervised data for fine-tuning is insufficient to achieve a high-quality text embedding model, likely due to its limited size. Conversely, unsupervised pre-training using web-scale text pairs yields superior text embeddings compared to solely relying on labeled data for fine-tuning. Nevertheless, the incorporation of supervised data in a multi-stage fashion with unsupervised pre-training can still contribute to the refinement of the acquired text embeddings.

\subsection{Training Data Mixture}
We study the influence of mixing ratio used in sampling distribution on pre-training data to model performance.

\begin{table}[!ht]
    \centering
    \begin{tabular}{cccc}
    \toprule
    $\alpha$ & Retrieval & STS & MTEB \\
    \midrule
    0 & 36.7 & 73.2 & 55.4 \\
    0.3 & 44.6 & 75.9 & 58.9 \\
    0.5 & 44.2 & 76.5 & 59.0 \\
    1 & 42.0 & 75.5 & 58.3 \\
    \bottomrule
    \end{tabular}
    \caption{Influence of ratio $\alpha$ used in pre-training data sampling.}
    \label{tab:alpha}
\end{table}

The performance on two task categories, retrieval and STS, as well as the average performance on MTEB is reported in Table~\ref{tab:alpha}.
We observe that neither uniformly sampling from each pre-training task ($\alpha=0$) nor directly combining all data sources ($\alpha=1$) is the best choice.
Setting $\alpha$ to 0.5 can improve results on all tasks.

\subsection{Ablation of the Contrastive Objective}

This work uses an improved contrastive objective which can efficiently enlarges the negative pool under fixed batch size.
We compare it against the vanilla contrastive loss with only in-batch negatives in both pre-training and fine-tuning stages.

\begin{table}[!ht]
    \centering
    \begin{tabular}{ccc}
    \toprule
    Setting & PT & FT \\
    \midrule
    Vanilla & 57.3 & 61.8 \\
    Improved & 57.8 & 62.4 \\
    \bottomrule
    \end{tabular}
    \caption{Comparison of the vanilla contrastive loss with in-batch negatives, and the improved contrastive loss with enlarged negative pool. For ablation we run the pre-training (PT) for 30k steps to reduce computational cost. We report average score on MTEB.}
    \label{tab:icl}
\end{table}

According to Table~\ref{tab:icl}, using improved contrastive loss consistently improves model performance in both pre-training and fine-tuning stages.

%% file: sections/discussion.tex
\section{Discussion}
Despite the strong performance on English tasks, our current model can only handle text with a length of less than 512, as it is initialized from BERT and lacks multilingual capabilities. Consequently, longer texts must be truncated or split for encoding. However, with more data engineering and compute resources, the described training approach could easily be extended to a multilingual version and accommodate longer contexts.

Another issue is the problem of data contamination resulting from large-scale pre-training on Internet data. Currently, we only conduct deduplication based on exact matching of text pairs, which is an overly strict filter. This issue has also been highlighted by ~\citet{gpt3} during the training of large-scale generative language models. We suspect that this is a common problem that other models also suffer from, but quantifying it without details about the training data sources is even more challenging~\citep{neelakantan2022text}.

Furthermore, the models trained in this study are based on a non-causal architecture with bidirectional context attention. It would be intriguing to explore similar pre-training methods for causal or prefix language models, as these models could optimize generation and retrieval jointly and unify them within a single model.

%% file: sections/appendix.tex
\section{More Details about Training Data}
\label{sec:appendix}

\subsection{Pre-training Data}

\paragraph{Web Page}
Within a web page, we use title as query and the body text as document.
Some resources include Common Crawl, Clue Webs, MS MARCO documents.
The task can be formatted as given a short title, find the most relevant body texts from a set of randomly sampled texts.

\paragraph{Academic Paper}
The scientific articles usually have a higher quality due to its formal nature.
For each paper, we use the title as query and its abstract as document for constructing text pairs.
The articles are mined from different websites (such as arXiv, bioRxiv, medRxiv, PubMed and Semantic Scholar) to cover a wide range of topics.

\paragraph{Hyperlink}
Another important information present on the internet is the hyperlink with text in it, also know as web anchors.
The hyperlink can provide necessary references for current arguments.
We use the citation argument and the text from reference as relevant text pairs for contrast.
This type of task is more challenging as it usually involves multi-hop reasoning.
We used three resources to incorporate the link information: ClueWeb, Wikipedia and Semantic Scholar paper citations.

\paragraph{Community QA}
We also used many data from community QA websites.
The UI design of such websites usually follows a structured format, where the user can write their questions in the format of a summaritive title and a descriptive body.
These two fields are usually semantically consistent.
In addition, we also consider the question answer pairs from this type of websites.
The data sources we used include StackExchange, Yahoo Answers, WikiHow and Amazon QA.
Simple heuristics such as text lengths and voting numbers are used to filter out low-quality data.

\paragraph{Social Media}
The social media websites such as Twitter and Reddit usually involves people publishing posts about one event, and many internauts leave their comments.
The post is also structured with title and body in it, which we consider as positive pairs.
Similar to Community QA, post comment are also regared as positive pairs for data mining.
We mine data from Reddit.

\paragraph{News}
News are structured as title body pairs.
Some news has highlighted sentences in it.
We use these information to construct (query,doc) pairs.
We used data from CCNews, MicrosoftNews, NPR, CNNDaily.

\paragraph{Knowledge Base}
Knowledge base usually stores textual descriptions knowledge about an entity or event.
The (entity, description) pairs are mined.
We use WikiPedia and DBPedia for text pair mining in this work.

\paragraph{Code}
Code can be viewed as another form of text.
The naturally paired text-code can be repurposed as positive pairs.
We use GitHub and StackOverflow as two data sources.
We reuse training set from CodeSearchNet which is mined from GitHub.

\paragraph{Others}
In addition, we also use data from various websites such as Amazon reviews about the goods, debate websites about one argument, googaq q,a pairs by prompting google search box with search log queries.

\subsection{Fine-tuning Data}

\paragraph{Web Search}
We used MS MARCO passage retrieval benchmarks.
Hard negatives are mined by sampling from high-ranked documents  retrieval system, excluding positive ones.

\paragraph{Open QA}
We consider Natural Questions, Trivia QA, Web Questions, HotpotQA, etc.
In the open domain QA datasets, a question and its supporting evidence passages are provided as positive pairs.
Top ranked passage by retrieval system which do not include answer to the question is regared as hard negatives.

\paragraph{Natural Language Inference}
Prior work~\citep{conneau-etal-2017-supervised} has shown that high-quality sentence embeddings can be learned from a supervised natural language inference task.
We use entailment as positive pairs and contradiction as negative pairs to construct training triples.
The combination of MNLI and SNLI is used in this work.

\paragraph{Fact Verification}
One argument and its supporting source (a Wikipedia document) is positive pairs.
We use training set from FEVER as data source for this task.

\paragraph{Paraphrase}
Two sentences with similar meanings are labeled as positive pairs.
This type of data includes Quora and StackExchangeDupquestion.

\paragraph{Others}
In addition to previous datasets, we also used miscellaneous datasets from different NLP tasks and domains released in MEDI~\citep{INSTRUCTOR} and BERRI~\citep{tart}.
By doing so, a sub-sampled version of pre-training data is also included in fine-tuning to avoid catastrophe forgetting.

\subsection{Data Sources}

The pre-training data comes mostly from language corpus released by previous work.
We use CommomCrawl preprocessed by CCNet at 2019 snapshot due to large computaional cost of processing~\citep{wenzek-etal-2020-ccnet}.
Since Reddit data is no longer free available, we use two pre-processed version by sentence-transformers \footnote{\url{https://huggingface.co/datasets/sentence-transformers/reddit-title-body}} and \citet{oguz-etal-2022-domain} for pair mining.
Text pairs mined from hyperlinks come from ~\citet{zhou-etal-2022-hyperlink} and \citet{xie23anchordr}.
We also include citation pairs from the S2ORC dataset~\citep{lo-wang-2020-s2orc}.
We reuse DBPedia, debating arguments and PubMed corpus from BEIR~\cite{beir}.
Wikipedia data is taken from~\citet{izacard_few-shot_2022}.
Microsoft News data comes from~\citet{wu-etal-2020-mind}.
Arxiv data is downloaded from Kaggle, medRxiv and bioRxiv are mined via requesting public API from year 2013 to 2022.
The StackExchange and StackOverflow data comes from the pre-processed version maintained by sentence-transformers team.\footnote{\url{https://huggingface.co/flax-sentence-embeddings}}
The remaining data comes from embedding-training-data.\footnote{\url{https://huggingface.co/datasets/sentence-transformers/embedding-training-data}}
The training data is keep as it was without any specific filtering, except that we use text pair exact-match for training data de-duplication for some datasets.

The fine-tuning data is basically a combination of previous research.
For the MS MARCO dataset, we use mined hard negative by the second stage retriever from~\citet{li2023cdmae}.
For NQ dataset, we reuse the training data released by coCondenser~\citep{gao-callan-2022-unsupervised}.
We use NLI data released by SimCSE~\cite{gao-etal-2021-simcse}.
Other data comes from MEDI and BERRI~\citep{INSTRUCTOR, tart}, but we discard the instructions written for each task and only use training triples.
Some randomly sampled examples can be found in Table~\ref{tab:ft_data}.

\input{tables/ft_data}

\section{Massive Text Embedding Benchmark}
\label{sec:mteb}
\paragraph{Classification}
This task is evaluated in the linear probing setting. The embedding model is kept frozen and used to extract text embeddings for each example from train and test set. The train set embeddings are used as input features to train a logistic regression classifier with 100 maximum iterations. The accuracy on test set is reported as the main evaluation metric. In this setting, different classification tasks only need to train an extra classification head with a few labeled training data.

\paragraph{Clustering}
A high-quality embedding model should embed semantically similar texts close in the embedding space. This property is evaluated by running a $k$-means algorithm on the embeddings produced for each sentence of the test set. A mini-batch $k$-means model is used with batch size 32 and $k$ being the number of labels. Texts are partitioned into $k$ clusters. The clustering performance is measured by the v-measure~\citep{rosenberg-hirschberg-2007-v} which is invariant to the permutation of clustering labels.

\paragraph{Reranking}
Given a query and a list of relevant and irrelevant reference texts, reranking needs to rank the list of reference texts based on their similarity to the query. The embedding model is invoked to obtain embeddings for each query and reference text and cosine similarity is used as the ranking score. This inference setting is quite similar to text retrieval with the reference set being smaller and harder to distinguish. In line with previous work, the main evaluation metric is MAP (mean average precision).

\paragraph{Retrieval}
We omit the text retrieval evaluation since it's similar to that introduced in previous section.

\paragraph{Pair Classification}
This task needs to assign a label for a pair of texts. Popular tasks include duplicate or paraphrase identification, where the label is binary. The similarity score is the cosine similarity between the embeddings of two texts. The average precision score is reported as the main evaluation metric using the best binary threshold.

\paragraph{Semantic Textual Similarity}
To determine the similarity between a given pair of sentences, continuous scores are assigned, with higher values indicating greater similarity. The embedding model is employed to embed the sentences, and their similarity is computed using cosine similarity. The estimated similarity scores is compared against human labeled scores ranging from 1 to 5. We report Spearman's correlation, which measures the rankings instead of the actual scores and better suits the need of evaluating sentence embeddings.

\paragraph{Summarization}
This is a text generation evaluation task which aims to automatically evaluate the quality of generated text.
For the summarization task, the quality of each generated summary is computed by measuring the cosine similarity between its embedding and the embedding of the ground truth references.
In the case of multiple gold references, the closest one with highest similarity score is used for quality estimation.
Similar to STS task, we use the Spearman correlation between the ranking produced by the text embedding model and the human assessments for evaluation.

\input{tables/codesearchnet}

\section{Original CodeSearchNet Results}
\label{app:csn}

We list the results of the original setting on CodeSearchNet in Table~\ref{table:code}, where the retrieval corpus contains 1k randomly sampled code snippets.
Compared to previous open-source code language models with similar architecture and size (CodeBERT~\citep{feng-etal-2020-codebert} and GraphCodeBERT~\citep{guo2021graphcodebert}), our model is superior in most programming languages.
There is still a performance gap to the code embedding model trained by~\citet{neelakantan2022text}, which used Codex~\citep{chen2021evaluating} as backbone and trained on a large-scale (code, text) pairs extracted from open-source code.
It is worthwhile to explore how to further close this gap.

%% file: tables/ft_data.tex
\begin{table*}[ht]
\centering
\resizebox{\textwidth}{!}{
\begin{tabular}{lllll}
\hline
Task Type & Text Triple Format & query & doc & hard neg \\ \hline
Web Search & \begin{tabular}[c]{@{}l@{}}(query, passage, negative)\end{tabular} & \begin{tabular}[c]{@{}l@{}}finger cellulitis symptoms\end{tabular} & \begin{tabular}[c]{@{}l@{}}The following are the most common \\ symptoms of cellulitis. However\ldots\end{tabular} & \begin{tabular}[c]{@{}l@{}}Cellulitis usually begins as \\ a small area of pain and \ldots\end{tabular} \\
\hline
Open QA  & (question, passage, negative) & \begin{tabular}[c]{@{}l@{}}big little lies season 2 \\ how many episodes\end{tabular} & \begin{tabular}[c]{@{}l@{}}Big Little Lies (TV series). \\ series garnered several accolades\ldots\end{tabular} & \begin{tabular}[c]{@{}l@{}}Little People, Big World. \\ final minutes of the season two\ldots\end{tabular} \\
\hline
Natural Language Inference  & (sentence, entailment, contradiction) & \begin{tabular}[c]{@{}l@{}}(Read  for Slate 's take \\ on Jackson's findings.)\end{tabular} & \begin{tabular}[c]{@{}l@{}}Slate had an opinion \\ on Jackson's findings.\end{tabular} & \begin{tabular}[c]{@{}l@{}}Slate did not hold any opinion \\ on Jackson's findings.\end{tabular} \\
\hline
Fact Verification &  \begin{tabular}[c]{@{}l@{}}(argument, evidence, others)\end{tabular}  & \begin{tabular}[c]{@{}l@{}}Roman Atwood is a \\ content creator.\end{tabular}    &  \begin{tabular}[c]{@{}l@{}}Roman Bernard Atwood (born \\ May 28, 1983) is an American \\ YouTube personality\ldots\end{tabular} & \begin{tabular}[c]{@{}l@{}}6th Streamy Awards Casey Neistat \\ and Jesse Wellens, PrankvsPrank \ldots\end{tabular} \\
\hline
Paraphrase &  \begin{tabular}[c]{@{}l@{}}(sentence, paraphrase, others)\end{tabular} & \begin{tabular}[c]{@{}l@{}}Lexapro taken with \\ crestor any reaction? \end{tabular} & \begin{tabular}[c]{@{}l@{}}Can dayquil be taken with Lexapro?\end{tabular} & \begin{tabular}[c]{@{}l@{}}Can stopping lexapro cause \\ a longer period?\end{tabular} \\
\hline
\end{tabular}
}

\caption{Examples of (query, positive, negative) text triples in fine-tuning data.}
\label{tab:ft_data}

\end{table*}

%% file: tables/codesearchnet.tex
\begin{table*}[!ht]
\setlength{\tabcolsep}{5pt}
\centering
\begin{tabular}{lcccccccc}
\toprule
Model & Params & Ruby & JS & Go & Python & Java & PHP & Avg. \\
\midrule
CodeBERT & 110M $\times$ 6 & 69.3 & 70.6 & 84.0 & 86.8 & 74.8 & 70.6 & 76.0 \\
GraphCodeBERT & 110M $\times$ 6 & 84.1 & 73.2 & 87.9 & 75.7 & 71.1 & 72.5 & 77.4 \\
\texttt{cpt-code} S & 300M & 86.3 & 86.0 & 97.7 & 99.8 & 94.0 & 96.7 & 93.4 \\
\texttt{cpt-code} M & 1.2B & 85.5 & 86.5 & 97.5 & 99.9 & 94.4 & 97.2 & 93.5 \\
\hline
GTE$_\text{base}$ & 110M & 79.6 & 79.4 & 84.2 & 98.8 & 86.8 & 86.8 & 85.9 \\
\bottomrule
\end{tabular}
\caption{
Results on CodeSearchNet~\cite{codesearchnet}.
We compare with CodeBERT~\cite{feng-etal-2020-codebert}, GraphCodeBERT~\cite{guo2021graphcodebert} and \texttt{cpt-code}~\citep{neelakantan2022text}.
This setting requires finding the relevant code block among 1K candidates for a given natural language query.
}
\label{table:code}
\end{table*}